\documentclass[conference]{IEEEtran}
\IEEEoverridecommandlockouts
\usepackage{cite}
\usepackage{amsmath,amssymb,amsfonts}
\usepackage{algorithmic}
\usepackage{graphicx}
\usepackage{textcomp}
\usepackage{cleveref}
\usepackage{attrib}
\usepackage{svg}
\usepackage{booktabs}
\newcommand{\EE}{\mathbb{E}}

\newcommand{\Xx}{ {\vec X = \vec x} }
\usepackage{mathtools, cuted}
\newcommand{\mcal}[1]{\mathcal{ #1}}
\usepackage{algorithm}
\usepackage{algorithmic}
\usepackage{siunitx}

\newtheorem{theorem}{Theorem}[section]
\newtheorem{definition}[theorem]{Definition}
\crefname{definition}{Definition}{Definitions}
\Crefname{definition}{Definition}{Definitions}

\def\BibTeX{{\rm B\kern-.05em{\sc i\kern-.025em b}\kern-.08em
		T\kern-.1667em\lower.7ex\hbox{E}\kern-.125emX}}
\begin{document}
	
	\title{Moral reinforcement learning using actual causation}

	\author{\IEEEauthorblockN{1\textsuperscript{st} Tue Herlau}
		\IEEEauthorblockA{\textit{DTU Compute} \\
			\textit{Technical University of Denmark}\\
			2800 Lyngby, Denmark \\
			tuhe@dtu.dk}
		}

	\maketitle
	
	\begin{abstract}
Reinforcement learning systems will to a greater and greater extent make decisions that significantly impact the well-being of humans, and it is therefore essential that these systems make decisions that conform to our expectations of morally good behavior. 
The morally good is often defined in causal terms, as in whether one's actions have in fact caused a particular outcome, and whether the outcome could have been anticipated. 
We propose an online reinforcement learning method that learns a policy under the constraint that the agent should not be the cause of harm. This is accomplished by defining cause using the theory of actual causation and assigning blame to the agent when its actions are the actual cause of an undesirable outcome.
We conduct experiments on a toy ethical dilemma in which a natural choice of reward function leads to clearly undesirable behavior, but our method learns a policy that avoids being the cause of harmful behavior, demonstrating the soundness of our approach. Allowing an agent to learn while observing causal moral distinctions such as blame, opens the possibility to learning policies that better conform to our moral judgments.
	\end{abstract}
	

	\section{Introduction}
Reinforcement learning deals with a problem where in time step $t$ the Agent observes a state, selects an action using a policy, and obtains a reward $r_{t+1}$ before transitioning to the next state. Current formulations of reinforcement learning determine the optimal policy as the one which maximizes the accumulated reward $\sum_{t=0}^{T-1} r_{t+1}$~\cite{sutton2018reinforcement}.  

It has been argued that reward maximization is enough to acquire behavior that mimics all known facets of natural intelligence, including motor, perception, social behavior and general intelligence~\cite{SILVER2021103535}. However, describing morally good behavior in terms of reward maximization is an extreme form of utilitarianism, which is known to lead to paradoxical behavior~\cite{foot1967problem,kagan1989limits}. 
	
Instead, what is morally (and often legally~\cite{pound2017introduction}) permissive is commonly described in causal language: was the agent's action a sufficient cause? Could the outcome be anticipated as a consequence of the agent's action?~\cite{oakley_cocking_1994,Sartorio2007-SARCAR}. Furthermore, cognitive science indicates that causality plays a crucial role in attribution of blame~\cite{gerstenberg2018lucky}. We hypothesize that in order to lead to morally acceptable behavior, the reward function of a reinforcement learning agent should be informed by these causal distinctions.
	
Recent work on \emph{actual causality} (see \cref{sec:ac}) shows exciting promise on this front, by providing operational meanings to morally significant terms such as cause, intent and blame, albeit in a non-temporal process, where the outcome of all relevant variables is known at once~\cite{halpern2005causes, halpern2018towards}.

In this paper, we propose a way to apply definitions of cause, found in actual causality, to the reinforcement learning setting, thereby letting the agent learn a policy which is informed by morally significant causal distinctions\footnote{Code to reproduce the results of this paper can be found at \texttt{https://gitlab.compute.dtu.dk/tuhe/moral\_agent/}. }.
	
We accomplish this by letting the states in reinforcement learning affect the exogenous variables in a relevant causal model and, based on the notion of actual cause derived from the model, issue a modified terminal reward signal to the agent: the relevant notion of cause is therefore described in the diagram, while the agent learns how and whether its policy affects the state of the causal model.

	The \emph{principle} we want the agent to follow is that \textbf{\emph{one is to blame for outcomes of events one causes, but not to blame for events one did not cause}}. Although by no means do we suggest that this provides a sufficient account of morality, it does touch upon key distinctions found in moral philosophy, such as cause and blame~\cite{Sartorio2007-SARCAR, hertzberg1975blame}. 
	
	\begin{figure}[t!]
		\centering
		\includegraphics[width=\linewidth]{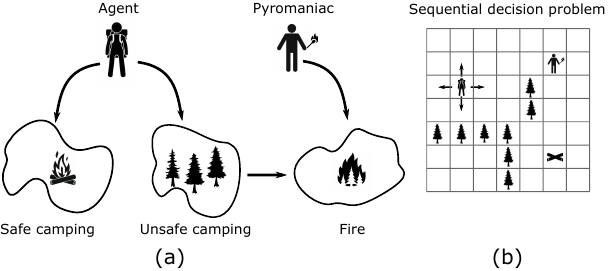}
		\caption{(a) Schematic illustration of the camping example. The agent can choose safe camping (reward of 10) or unsafe, easier camping (reward of 20). The forest burning down, which occurs under unsafe camping, results in a reward of -100. A pyromaniac is also at large, and will always burn down the forest. (b) Sketch of the camping example as a sequential decision problem. The problem now includes several time steps, but has the same structure.}\label{fig1}
	\end{figure}

	Philosophers and experimental psychologists commonly illustrate, test and challenge our moral intuition using vignettes such as the trolley problem~\cite{clifford2015moral,foot1967problem}, and we will similarly illustrate and test our method using a vignette (camping in a dry forest, \cref{fig1} (a)), which is inspired by~\cite{paul2013causation, halpern2005causes}.

	\begin{quotation}
		Albert is camping and wants to start a fire. Albert can set up camp at either a safe or unsafe location. The unsafe location is easier to get to, but it will also cause the forest to burn down. A pyromaniac is on the loose, and will always burn down the forest. 
		\attrib{Camping in a dry forest}
	\end{quotation}

	The vignette illustrates a problem of simple consequentialism, namely that although the \emph{consequence} of Albert's two options (camping safely and unsafely) is the same (the forest will burn down), Albert's decision to camp at an unsafe location \emph{still deserves blame}: what matters in the example is whether Albert \emph{caused} the forest to burn~\cite{paul2013causation}.

	The above vignette can be considered a reinforcement learning problem with the same conclusion (see \cref{fig1} (b)), where the environment is described using states $s_t$ (the state could contain the agent's state, the pyromaniac's actions, and so on), and actions $a_t$. We assume a natural reward structure in which the agent obtains a reward $r=10$ for camping at the safe camping spot, $r=20$ for unsafe camping spot (camping unsafely is less strenuous) and finally, a reward of $r=-100$ for the bad outcome of the forest burning down. In this scenario, the pyromaniac eventually burns down the forest, and therefore the possible accumulated reward the agent will obtain is either $-100$ (no camping), $r=-90$ (safe camping) and $r=-80$ (unsafe camping). An agent which maximizes the reward will therefore learn to always camp unsafely. 
	
	The point of the example is not to argue that the behavior cannot be avoided by suitable manipulations of the reward signal (indeed, this is exactly what our approach does). Rather, we suggest that this modification of the reward signal should be grounded in causal analysis, using a relevant causal diagram (\cref{fig1b}) and an objective definition of cause~\cite{halpern2005causes}. The agent must then learn which causal variables in the diagram it can manipulate and therefore may be blamed for (see \cref{sec:blame}), and this will form the basis for the modified reward signal. When a variable is cause for a bad outcome, the agent obtains a negative reward, but otherwise it does not obtain a negative reward, since it is not to blame. For this reason, we have dubbed the method \emph{morally aware reinforcement learning}.

	The contributions of this paper are therefore threefold: 
	\begin{itemize}
		\item We use an example to illustrate how reward maximization can lead to morally impermissible behavior in a concrete reinforcement learning problem.
		\item We propose a general variant of $Q$-learning which uses causal information to modify the reward signal so as to avoid blameworthy actions.
		\item We use experiments to demonstrate how this method can avoid morally impermissible behavior 
	\end{itemize}
	
	\subsection{Related work}
	The problem of assigning blame in a sequential, team-plan setting was considered in \cite{alechina2020causality}, where structural equations and actual causality were used to assign blame to each agent for its contribution to the overall outcome. However, this work differs from ours in that it does not consider learning, and is concerned with the combined actions of a team. 
	
	Several other references consider a combination of reinforcement learning and ethics/morality in which ethics is introduced as a weighting factor in the reward function. For instance, \cite{ecoffet2021reinforcement} consider how different weighting procedures allow moral theories (functions) to vote for particular behaviors, but otherwise leave the specification up to a designer. Other approaches specify these functions through learning from (external) demonstrations~\cite{noothigattu2019teaching} or from queries to an idealized (hidden) ethical utility function~\cite{abel2016reinforcement}, or supervised using text-based data~\cite{hendrycksmoral} in a text-based game. Although the objective is similar to ours, the above work differs in that the ethical theories are \emph{externally} specified to the agent, whereas our procedure has a built-in notion of blame, which it combines with a learning procedure of which variables may be influenced.

	In  \cite{kleiman2015inference},  actual causation is used as a way to \emph{explain} human judgments about intention (but with a different definition of intent than \cite{halpern2018towards}) in problems with a fixed number of states and decisions. This approach differs from ours in that the goal is to explain judgments, and not to learn a policy which conforms to limitations imposed by the definitions of intent.

	\section{Methods}\label{sec:methods}
	Causation can be understood in two ways: as \emph{general statistical causality} (i.e., does smoking cause cancer?) or as \emph{actual causality} which focuses on specific events (did Bob's smoking cause his lung-cancer?)~\cite{halpern2005causes,halpern2016actual}. Clearly, the latter form of causality is the more relevant one when it comes to moral dilemmas, since these should always be decided with a specific situation in mind~\cite{kleiman2015inference}.   
	
	Both types of causality are mathematically expressed using variations of structural causal models (SCMs)~\cite{pearl2009causality}, and we will briefly review the main definitions below as they relate to the Camping example.
	
	\begin{figure}[t]
		\centering
		\includegraphics[width=.8\linewidth]{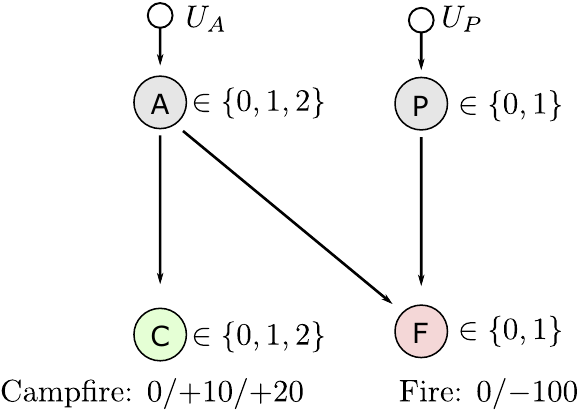}
		\caption{The camping example abstracted into a SCM. When described as an SCM, the variables are non-temporal and assumed known all at once. Based on the graph structure, actual causation allows us to define whether certain variables are the cause of others. Large circles are endogenous variables $\mcal V$, and the small circles are the exogenous variables $\mcal U$}\label{fig1b}
	\end{figure}
	
	\subsection{Actual causation} \label{sec:ac}
	Our discussion of actual causation and SCMs will closely follow \cite{halpern2005causes}. 
	Actual causation assumes that there exists a relevant model of the world described in terms of variables, their values and how they influence each other. Variables are classified as either \emph{exogenous variables}, whose values are determined by factors outside the causal model itself, or \emph{endogenous variables}, whose values are determined by the exogenous variables. Specifically, a \emph{structural causal model} $M$ is a tuple $(\mcal S, \mcal F)$. $\mcal F$ is a set of \emph{structural equations}, which denote a deterministic relationship between the variables, and $\mcal S$ is the \emph{signature} consisting of the triplet $(\mcal U, \mcal V, \mcal R)$, where $\mcal U$, $\mcal V$ are the exogenous/endogenous variables respectively, and for any variable $Y \in \mcal U \cup \mcal V$ then $\mcal R(Y)$ is the range of values the $Y$ can take. 
	
	In the case of the camping example, the endogenous variables are discrete variables such as Fire $F$ ($F=0$ if there is no fire and otherwise $F=1$), $A=0$ (no action), $A=1$ (set up campfire at a safe location) and $A=2$ (set up campfire at an unsafe location). The pyromaniac's actions are $P=0, 1$, depending on whether the pyromaniac starts a fire. Finally, we include the variable $C=0$ (no campfire), $C=1$ (difficult to set up campfire) and $C=2$ (easy to set up campfire) in the model, for clarity, although it is not strictly required to analyze the example.

	The exogenous variables $U_A$ and $U_B$ capture the intentions/actions of the agent and the pyromaniac, see \cref{fig1b}. The overall example is an extension of the forest fire example in \cite{halpern2005causes}, with the main difference being explicitly defining $C$ and including the no-camping option $A=0$, which will be required for reinforcement learning.

	For each endogenous variable $X \in \mathcal{V}$ there is a function $F_X \in \mathcal F$ such that $F_{X}:\left(\times_{U \in \mathcal{U}} \mathcal{R}(U)\right) \times\left(\times_{Y \in \mathcal{V}-\{X\}} \mathcal{R}(Y)\right) \rightarrow \mathcal{R}(X)$. That is, $F_X$ determines the value of $X$, given the other variables in $\mcal U \cup \mcal V$.

	We limit ourselves to acyclic models and the model is commonly drawn as a directed graph, where $\mcal U \cup \mcal V$ are the vertices and there is a directed arrow from $Y$ to $X$, provided that $F_X$ has $Y$ as an input argument. 
	
	In the camping scenario, these functions are particularly simple, for instance $a = F_A(u_A) = u_A$ and $f = F_F(a, p) = \max\{\delta_{a,2},p\}$ (realizations of variables are written in lower case). 
	
	An assignment of value to all exogenous variables $\mcal U$ is called a \emph{context} and is denoted $\vec{u}$ (in our case simply $\vec u = (u^A, u^P)$), and the context will uniquely define the value of all endogenous variables. 
	
	\subsection{Interventions and causation}
	Given a causal model, the effect of external interventions is defined as follows: suppose we set a variable $X \in \mathcal{V}$ to $x$, the result is a new causal model, denoted $M_{X \leftarrow x}$, which is identical to $M$, except the equation $F_X$ is replaced by the constant $F_X = x$. We can generalize this idea to define effects of interventions as follows:
	
	In a causal model $M= (\mcal S, \mcal F)$ with signature $\mcal S = (\mcal U, \mcal V, \mcal R)$, a \emph{primitive event} is a formula of the form $X = x$ for $X \in \mathcal{V}$ and $x \in \mathcal{R}(X)$. A \emph{causal formula} is of the form $\left[Y_{1} \leftarrow y_{1}, \ldots, Y_{k} \leftarrow y_{k}\right] \varphi$, where
	\begin{itemize}
		\item $\varphi$ is a Boolean combination of primitive events 
		\item $Y_1,\dots,Y_k$ are distinct variables in $\mcal V$
		\item $y_i \in \mathcal{R}(Y_i)$ is a legal value of $Y_i$.
	\end{itemize}
	The intuitive meaning of a causal formula is that $\varphi$ would hold (be true) if $Y_i$ was assigned to $y_i$ for all $i=1,\dots,k$.  
	A causal formula is abbreviated as $[ \vec{Y} = \vec{y} ]_\varphi$ and the special case $k=0$ as just $\varphi$. Whether a causal formula $\varphi$ is true or false in a causal model depends on the context. 
	We write $(M, \vec{u}) \models X=x$ if the variable $X$ has value $x$ in the context $\vec{u}$. The truth of conjunction and negation of primitive events can be defined analogous, and we can define the relation $\models$ by letting $(M, \vec{u}) \models [\vec Y = \vec y]_\varphi$ provided that $(M_{\vec{Y}=\vec{y}}, \vec{u} ) \models \varphi$, where $\varphi$ is any Boolean combination of primitive events. 
	Given these definitions, \emph{cause} can be defined as follows~\cite{halpern2005causes}: 
	\begin{definition}\label{def:cause}
		$\vec X = \vec x$ is an \emph{actual cause} of $\varphi$ in $(M, \vec u)$ provided that
		\begin{description}
			\item[(AC1)] $(M, \vec{u}) \models(\vec{X}=\vec{x})$ and $(M, \vec{u}) \models \varphi$	
			\item[(AC2a)] $\, $ There is a partition of the endogenous variables $\mcal V$ into two disjoint subsets $\vec W$ and $\vec Z$ such that $\vec X \subset \vec Z$ and a setting $\vec x'$ and $\vec w'$ of $\vec X$ and $\vec W$ such that 
			\begin{align}
				(M, \vec{u}) \models [\vec{X} \leftarrow \vec{x}^{\prime}, \vec{W} \leftarrow \vec{w} ]_{\neg \varphi}.
			\end{align}		 
			\item[(AC2b)] $\, $ If $\vec z$ is such that $(M, \vec u) \models \vec Z = \vec z$, then, for all subsets $Z'$ of $\vec Z$
			\begin{align}
				(M, \vec{u}) \models [\vec{X} \leftarrow \vec{x}, \vec{W} \leftarrow \vec{w}, \vec Z' \leftarrow z' ]_{\varphi}.
			\end{align}		
			\item[(AC3)] No subset of variables in $\vec X$ satisfy condition AC1 and AC2.		
		\end{description}
	\end{definition}
	This definition is quite technical, and it is best understood through examples~\cite{halpern2005causes}. Here, we will limit ourselves to the campfire example. Suppose we want to say unsafe camping $\vec X = \{A\} = \{2\}$ is a cause of fire $\varphi = F=1$, even when the pyromaniac is on the loose $P= 1$. AC3 says our cause should be minimal (trivially true since it is a singleton), and AC1 says that both $A=2$ and $F=1$ had to occur for $A=2$ to be considered a cause of $F=1$.

	It is AC2 which does the hard work. For $\vec X=\vec x$ to be a cause, we want to say that if $\vec x$ had not occurred, then the effect ($\varphi = (F=1)$) would not occur either (a basic but-for condition). But since $P=1$ only ensures $F=1$, the definition must introduce a contingency $\vec W = \vec w$ under which $F\neq 1$; In our example, $\vec W = \{P\} = \{0\}$ will satisfy AC2a. 
	
	Since this loosens the condition for something to be a cause, AC2b is required to rule out certain counterexamples~\cite{halpern2005causes}, however in our example $\vec Z = \{C, F, A\}$ is irrelevant since $\vec X = A = 2$ alone is sufficient to ensure $F=1$. As we can see, in the scenario, unsafe camping $A=2$ (and by symmetry, the pyromaniac setting the forest alight, $P=1$) is considered a cause of fire $F=1$, even though if any one of them did not occur, the forest would still burn.

	It should be mentioned that other definitions of cause are possible (see e.g. \cite{halpern2015modification}), however, the practical differences are minute, and the following discussion applies equally to such modifications of \cref{def:cause}.
	
	\subsection{Overview of our approach} \label{sec2.3}
	We will consider a standard episodic reinforcement learning setup where the environment evolves between states $s_t$ over times steps $t=0,\dots,T$. The agent takes action $a_t$ and, upon transition to state $s_{t+1}$, obtains a reward $r_{t+1}$. The states/actions/rewards are assumed to follow a \emph{Markov Decision Process}\cite{sutton2018reinforcement}. 
	
	The crucial point is how the states, actions and rewards, which occur over time, are connected to causation as defined on the non-temporal SCM as discussed in \cref{sec:ac}. Our assumptions will be similar to those in other work on structuring reward signals, such as \emph{reward machines}~\cite{icarte2020reward}. That is, we assume that the basic structure of endogenous/exogenous nodes of the SCM is a given, and that it is known how the states $s_t$ affect the value of the exogenous nodes $\mcal U$. As is the case in reward machines for reinforcement learning, this type of identification between states and course-grained variables is often natural to define~\cite{icarte2020reward}, and we do not make assumptions that the outcome of the \emph{policy} is known.

	The identification between states and \emph{exogenous} variables is also natural from the point of actual causation, where the role of the exogenous variables is exactly to model all factors affecting the assignment of the endogenous variables, like chance events, motives and other contingent factors (see discussion in \cite{halpern2005causes}).

	To deal with the temporal aspect, we observe that in other causal vignettes such as the original forest scenario~\cite{halpern2018towards}, or the trolley-cart scenario~\cite{clifford2015moral}, there is an implicit assumption that the exogenous variables $\mcal U$ \emph{obtain} their value \emph{during} some temporal process (examples include a trolley-cart driving down a track and hitting a person, a person setting a forest alight and so on), and once this process has assigned them their value they do not change. This suggests a  process where for each $U_k \in \mcal U$, we define the temporal (assigned) value as an absorbing process $u_i(t), t=-1,\dots, T$ as
	\begin{align}
		u_k(t) & =\begin{cases} u_k(t-1) & \text{if } u_k(t-1) \neq u_k(-1) \\
			h_{u_k}(s_t)   & \text{otherwise }      %
		\end{cases}\label{eqn:3}
	\end{align}
	That is, an exogenous variable $U_k \in \mcal U$ is given a default value $u_k(-1)$, and can then possibly be assigned a different value $h_{u_k}(s_t)$ at a later point, but after this assignment has occurred it will not change. The actual value of the exogenous variables is then defined as $U_k = u_k = u_k(T)$. In our example, $u_k(-1) = 0$ will be used as the natural default value. Once assigned, we can compute the value of all endogenous values and use these to compute the reward signal.

	Crucially, this assignment of credit must take into account whether the agent was responsible for the negative outcome (in this case the fire $F$).  As discussed in the introduction, the agent is considered responsible exactly if the agent's actions were an actual cause of the event. Therefore, when a particular event $F$ occurs, we use \cref{def:cause} to compute all actual causes of the event $\vec X^F_1, \dots, \vec X^F_k$.  In the camping example, these can be $A=2$ and $P=1$. For each cause $\vec X_k$, we compute the degree to which the agent is blameworthy of the cause $B_{\vec X_k}$ (defined in \cref{sec:blame}) and finally let the reward signal be the maximum blameworthiness of the causes of the outcome -- the intuition being that if the agent is fully responsible for just one true cause of the outcome, the agent is responsible for the outcome. The contribution to the terminal reward $r_T$ is therefore: 
	\begin{align}
		\max_{\mbox{$\vec X_k$ actual cause of $F$}} B_{\vec X_k} r( F) \label{eq:reward0}
	\end{align}
	where $r$ is a reward function affecting the exogenous variables, in our case simply $r(F) = -100 F$. 
	
	\subsection{Blame and manipulation}  \label{sec:blame}
	\begin{figure}[t!]
		\centering
		\includegraphics[width=.8\linewidth]{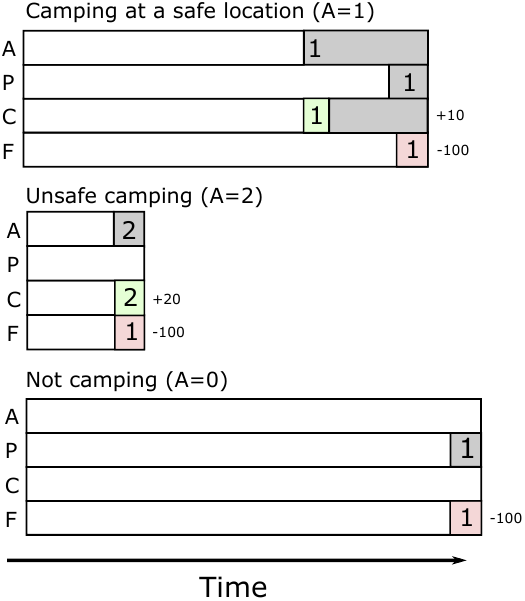}
		\caption{Temporal evolution of the relevant variables from the Camping example and the obtained reward (see \cref{fig1}). Note that unsafe camping makes it appear as if the agent can manipulate the pyromaniac's actions $P=1$, because the environment terminates before $P=1$ is observed.  }\label{fig2}
	\end{figure}
	
	In order to assign blame to the agent for outcomes $\Xx$, for instance $A=2$ or $P=1$, the agent must be able to manipulate the events through its actions. This basic definition of blame can be made more detailed to account for trade-offs in rewards/costs~\cite{halpern2018towards}. However, for simplicity we will use the simpler manipulation criteria, which will suffice for our example, and illustrate the main difficulties in applying the definition to reinforcement learning. 

	The main obstacle in defining blame is that the probability of an event occurring (or not) under different policy choices $\pi$, $P(P=1|\pi)$, does not by itself capture important aspects of whether an event can be manipulated or not (see \cref{fig2}). For instance, suppose we know that safe camping, given enough time, will eventually burn down the forest, it will still be true that unsafe camping, by hastening the fire, is a blameworthy action. At the same time, whether the pyromaniac burns down the forest may \emph{appear} within our control, because we can \emph{choose} to burn down the forest before the pyromaniac is given a chance (\cref{fig2}, bottom), so our definition should exclude this case.

	Our criteria for when a variable can be manipulated will therefore focus on whether the agent's actions \emph{reduces} the (estimated) number of steps $T_{\Xx}^{(\eta)}(s_t, a_t)$ until $\Xx$ \emph{is expected to occur}, given the agent is in state $s_t$ and takes action $a_t$. The parameter $\eta$ measures how conservative the estimate is. We will return to the practical estimation of $T_{\Xx}^{(\eta)}(s_t, a_t)$ in  \cref{sec:T}, but for now we assume it is given, and focus on how it can be used to define blame.

	Intuitively, blame for $\Xx$ increases when we take actions which \emph{decrease} the time until $\Xx$ occurs, relative to actions which seek to postpone $\Xx$. A tentative definition of blame is therefore: 
	\begin{align}
		1 - \frac{ T^{(\eta)}_{\Xx}(s_t, a_t)  }{  T^{(-\eta)}_{\Xx}(s_t) }, \ 
		T^{(\eta)}_{\Xx}(s_t) = \max_{a'} T^{(\eta)}_{\Xx}(s_t, a') \label{eq:8a}
	\end{align}
	That is, simply the relative reduction in time compared to the action $a'$, which seeks to maximally postpone $\Xx$. However, two situations should be taken into consideration:
	
	\begin{enumerate}
		\item A potentially catastrophic action and its effect may be separated by several steps; furthermore, if the intermediate actions are optimal, this does not reduce the blame for the catastrophic action.
		\item Blame is not simply cumulative over time: a driver who normally drives recklessly, but is involved in an accident on a day that he/she happens to drive safely, should not incur blame.
	\end{enumerate}
	Both factors can be accounted for by tracking the longest known time until $\Xx$ under actions that seek to \emph{delay} $\Xx$ as much as possible:
	
	\begin{align} \label{eq:8}
		T^{(\eta),+}_{\Xx} (s_t)  = \begin{cases}
			T^{(\eta)}_{\Xx}(s_t) & \mbox{if $t=0$} \\
			\max\left\{T^{(\eta),+}_{U=u}(s_{t-1} ) -1,  T^{(\eta)}_{U=u}(s_t)  \right\} 
			& \\
		\end{cases} 
	\end{align}
	and then define blame analogous to \cref{eq:8a} as:
	\begin{align}
		B_{\vec X=\vec x}(s_t, a_t) = 1 - \frac{ T^{(\eta)}_\Xx(s_t, a_t)  }{ T^{(-\eta),+}_\Xx  (s_t)   }. \label{eq10}
	\end{align}
	The maximum in \cref{eq:8} ensures this quantity is always between $0$ and $1$. The definition accounts for the two aforementioned situations as follows
	\begin{itemize}
		\item Suppose that at an earlier time step $t$ the time until $\Xx$, $T^{(\eta)}_{\Xx}$, is very large and that a bad action $a_t$ reduces this time. Although subsequent actions by the agent are optimal, the left argument of the maximum in \cref{eq:8} will ensure that $T^{(\eta),+}_{\Xx}$ remains large in subsequent steps, thus ensuring that the agent is still blamed for the earlier (bad) action in \cref{eq10}.
		\item On the other hand, suppose that the agent has taken several bad actions, thereby resulting in a blameworthy state according to \cref{eq10}. If a good action brings the agent to a state $s_t$ which is safe, i.e. $T^{(\eta)}_{\Xx}$ is large, then \cref{eq:8} (right-hand part of the maximum) will reset the maximum time $T^{(\eta),+}_{\Xx}$ to a large value, and the agent will no longer be blamed. 
	\end{itemize}

	\begin{algorithm}[t] 
		\hspace*{\algorithmicindent} \textbf{Input} SCM $M$ with endogenous variable $F$ used in reward function \cref{eq:reward0} \\
		\hspace*{\algorithmicindent} \textbf{Output} Policy $\pi$ which maximizes reward while avoiding blameworthy actions 
		\caption{Proposed method}\label{alg1}
		\begin{algorithmic}[1]
			\STATE $L \gets $ Empty list of potential actual causes
			\STATE Initialize $Q$-values defining the $\varepsilon$-greedy policy $\pi$
			\FOR{Each episode}		
			\STATE Reset to state $s_0$ and compute first action $a_0$
			\FOR{Each $t=0,\dots,T$}
			\STATE Use action $a_t$ and obtain $r_{t+1}, s_{t+1}$ from environment
			\IF{$s_{t+1}$ is not terminal}
			\STATE Compute $a_{t+1}$ using $\pi$
			\FOR{Each potential AC $(\Xx) \in L$}
			\STATE Update $p_{s_t,u_t} = P(\Xx | s_t, a_t)$ by tracking the sample average 
			\STATE Update the corresponding moment estimates $m_1$, $m_2$ using \cref{eq:m1m2}
			\STATE Update $T^{ (\eta), +}_{\Xx}$ using \cref{eq:8}
			\ENDFOR
			\ELSE		
			\STATE If $F \neq 0$, compute all actual causes $\vec X^{k}=\vec x^k$ of $F$ using \cref{def:cause}
			\STATE Compute blame using \cref{eq10} and add modified reward \cref{eq:reward0} to terminal reward $r_T$ 
			\STATE Add any previously unseen causes $(\vec X^k=\vec x)$ to $L$
			\ENDIF	
			\STATE Train $Q(s_t,a_t)$ using Q-learning and $r_{t+1}$ 
			\ENDFOR	
			\ENDFOR
		\end{algorithmic}
	\end{algorithm}

	\subsection{Estimating temporal delays}\label{sec:T}
	To estimate $T^{(\eta)}_{\Xx}(s_t,a_t)$, the expected number of steps until $\Xx$, assuming that the agent starts in state $s_t$ and takes action $a_t$, we will use an iterative method similar to $Q$-learning. To see how, first note that a simple application of the rules of probability shows that the time until an event $\Xx$, denoted by $T_\Xx$, follows the distribution: 
	\begin{align} \label{eq:main}
		& P_\pi(T_{\Xx} = t | s_t, a_t) =
		\delta_{0}(t) P(\Xx  | s_t, a_t)  + \\
		& 
		(1-\delta_{0,t} )
		(1-P( \Xx | s_t, a_t) ) 
		\EE_{s_{t+1},\pi}\left[T_{\Xx}-1 | s_{t+1} \right], \nonumber  
	\end{align}
	where $s_{t+1}$ is the next state in the MDP and $P(\Xx|s_t, u_t)$ is the chance $\Xx$ in this very same time step. The expectation is taken with respect to $s_{t+1}$ and $a_{t+1}$ as generated by the MDP and policy $\pi$. For terminal steps $s_{t+1}$, the estimate $P_\pi(T_{\Xx} | s_{t+1})$ on the right-hand side is replaced by a dispersed (normal) prior\footnote{In the experiments, the mean and variance is set to 10, and the method was robust to different choices.}, reflecting our belief of when $T_{\Xx}$ occurs after the episode has terminated. 
	
	This expression is computationally inefficient, as it would require us to keep track of the probability of each value of $T_\Xx$. However, this would also be unnecessary, since what matters is the mean time until the event occurs and not the overall shape of the distribution. We will therefore use a normal approximation so that: $T_{\Xx } | s_t, a_t \sim  \mcal{N}( m_1(s_t, a_t),  \sigma(s_t, a_t) )$ and simply track how the mean/variance is updated. 
	
	\begin{figure}[t!]
		\includegraphics[width=\linewidth]{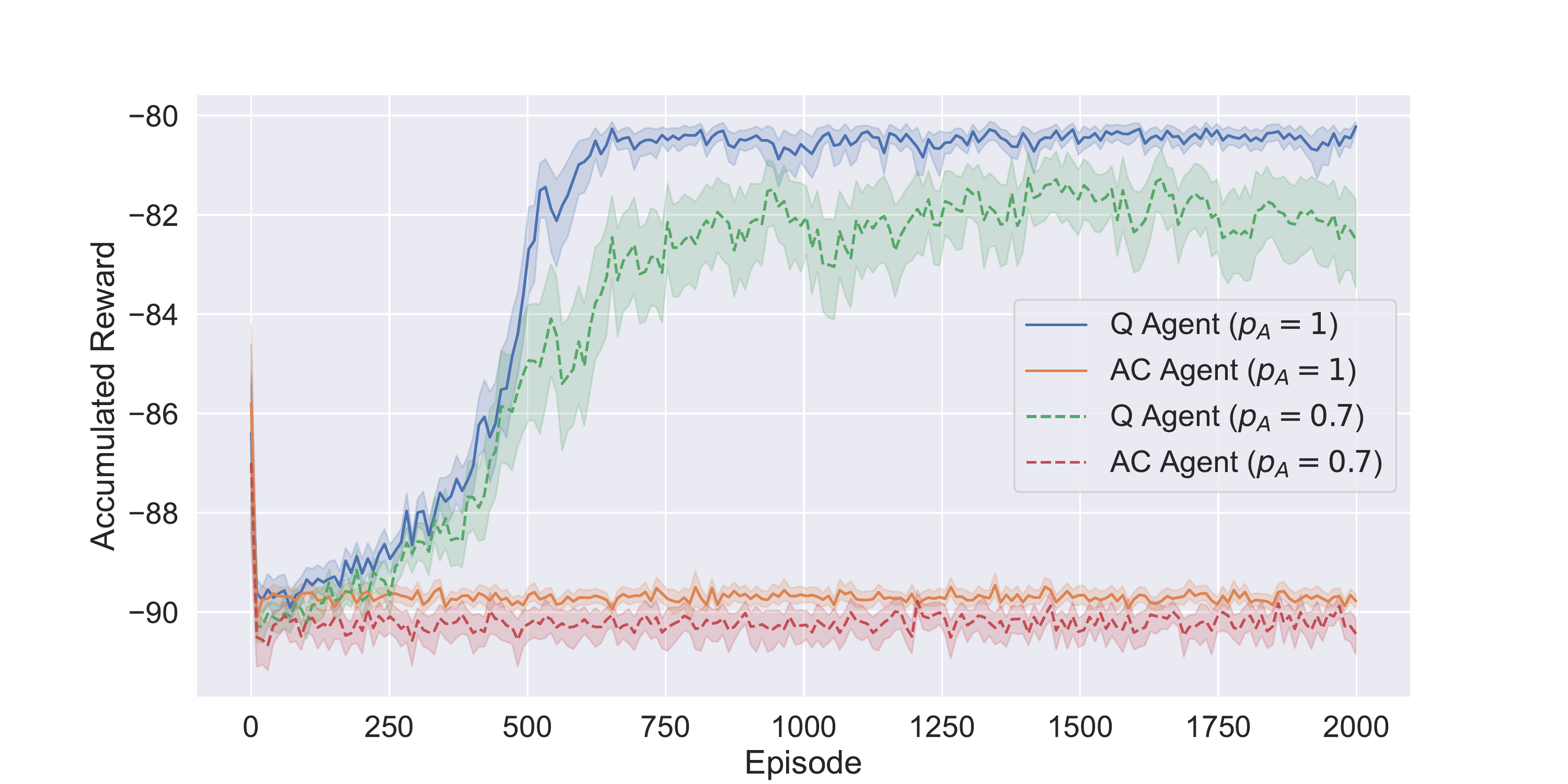}
		\caption{Traceplot of accumulated return per episode for a Q-learning agent and our actual cause method \cref{alg1} (AC Agent). The responsible behavior in the example gives a reward of $-90$, which our method obtains. Meanwhile, the Q-learning agent learns to burn down the forest.}\label{trace}
	\end{figure}

	This allows us to simplify \cref{eq:main} by plugging in the normal distribution and performing matching. Using the shorthand $p_{s_t,u_t} = P(\Xx|s_t,a_t)$ (this value is easily estimated as the sample average) we obtain the following expressions for the first and second moments:
	\begin{subequations}
		\begin{align}
			m_1 & = (1-p_{s_t, u_t} )(1+ \EE m^{(1)}_{s_{t+1}, a_{t+1} } ) \\
			m_2 & = (1-p_{s_t, u_t} )\left(\EE \sigma^2_{s_{t+1}, a_{t+1} } + (1+\EE m^{(1)}_{s_{t+1}, a_{t+1} } )^2  \right).
		\end{align}
	\end{subequations}
	Since the right-hand side only involves simple expectations, we can obtain asymptotically convergent estimates iteratively using $\alpha$-soft updates as in $Q$-learning. The updates are\footnote{The symbol $x \leftarrow y$ is understood either as a running average of all $y$-values or $\alpha$-soft updates $x = x + \alpha (x-y)$.}:
	
	\begin{subequations}
		\begin{align}
			p_{s_t,a_t} & \leftarrow 1_{\Xx} \\
			m^1_{s_t,a_t} & \leftarrow (1-p_{s_t, a_t} )(1 + m^1_{s_{t+1}, a_{t+1} } ) \\
			m^2_{s_t,a_t} & \leftarrow 
			(1-p_{s_t, a_t} )(1+ m^2_{ s_{t+1}, a_{t+1} } + 2 m^1_{s_{t+1}, a_{t+1} } )
		\end{align}\label{eq:m1m2}
	\end{subequations}

	Since the time estimates are uncertain, we define $T^{(\eta)}_{\Xx}(s_t, a_t) = m^{(1)}_{s_t, a_t} + \eta \sigma_{s_t, a_t}$ as the $\frac{1}{2} + \eta$ percentile of the estimated number of steps until $\Xx$ occurs in state $s_t$ under action $a_t$. Together with the definition of blame in \cref{sec:blame} we can compute the modified reward signal in \cref{eq:reward0}, and the full method can be found in \cref{alg1}.

	\begin{table}[t]
		\centering
		\caption{Average return without exploration} \label{tbl3}	
\begin{tabular}{ lcc }
\toprule Condition
 &  AC Agent
 &  Q Agent
   \\ \midrule $p_A = 1$
 &  \SI{-90.008\pm 0.004}{}
 &  \SI{-80.000\pm 0.000}{}
   \\  $p_A = 0.7$
 &  \SI{-90.640\pm 0.193}{}
 &  \SI{-82.286\pm 0.377}{}
   \\ \bottomrule\end{tabular}
	\end{table}
	\section{Experiments} 
	We test the proposed method (\emph{AC Agent}) on a MDP variant of the camping vignette. To avoid unnecessary details, we will consider a case where the states and actions are similar to the deterministic case shown in \cref{fig1b}, but where the effect of actions is not deterministic, so that the interaction with the environment will occur over an (unknown) number of time steps. We assume that there is a chance $p_\text{pyro} =0.1$ that the pyromaniac will set the forest alight in each step, and with probability $p_A$ the agent's actions will have no effect (and otherwise have the effect of camping safely and unsafely). We compare the method against tabular $Q$-learning~\cite{sutton2018reinforcement}. For both methods, we set the learning rate to $\alpha =0.05$, the $\varepsilon$-greedy exploration rate to $\varepsilon = 0.1$, and the discount factor to $\gamma=0.99$. We selected $\eta=0$ in the definition of blame. The choice of these parameters is not important, and only affects speed of convergence; the full code listing is available online.

	We first examine the convergence of the methods by plotting the accumulated return per episode during training. The result can be found in \cref{trace} (using $2000$ steps per episode. The plot is the average over $50$ restarts, and the shaded regions indicate the standard deviation of the mean). We see that the $Q$-agent converges to the optimal policy (to always burn down the forest) which, as discussed in the introduction, is associated with a higher reward. The AC agent chooses a responsible policy (safe camping) and therefore obtains a lower reward. Although the results are noisy due to the $\epsilon$-greedy exploration, we verify that the methods indeed converge by testing the 50 trained agents on 100 episodes each, with exploration. The average return is shown in \cref{tbl3}, and indicates that the only variation occurs when the pyromaniac burns down the forest before the agent has a chance of performing safe camping. 
	
	To verify that this indeed occurs, because the agent attributes blame correctly, we extract the blame factor (\cref{eq:reward0}) for the two actual causes of $F$, namely the agent setting up an unsafe campfire ($A = 2$) and the pyromaniac setting the forest alight, $P=1$ (same settings as above). The result can be found in \cref{tbl:blame}, and we see the agent correctly identify the pyromaniac's actions as being outside its control (i.e. a blame of 0 when they are an actual cause of the forest burning down), while its own choice of unsafe camping has a high degree of blame. That this factor is not identical to 1 is due to blame being computed as a fraction of reduced time until an event occurs, and since the estimated times are all finite, this naturally places an upper limit on the degree of blame (compared to a similar choice in the degree of intent, as defined in \cite{halpern2018towards}).

	\begin{table}[t]
		\centering
		\caption{Estimated blame factors}\label{tbl:blame}	 
\begin{tabular}{ lcc }
\toprule 
 &  Pyromaniac: $B_{P=1}$
 &  Unsafe campfire: $B_{A=2}$
   \\ \midrule $p_A = 1$
 &  \SI{0.002\pm 0.002}{}
 &  \SI{0.989\pm 0.000}{}
   \\  $p_A = 0.7$
 &  \SI{0.016\pm 0.009}{}
 &  \SI{0.741\pm 0.017}{}
   \\ \bottomrule\end{tabular}
	\end{table}

	\section{Conclusion}
	In this paper, we have proposed a method which allows a reinforcement learning agent to distinguish between morally good and bad behavior, namely that the agent is behaving badly if its actions are an actual cause of a bad outcome, so that even if a bad outcome is unavoidable, the agent should not be responsible for it in any case. 
	
	We have formalized this using a definition of cause taken from the theory of actual causality, and our main contribution has 
	been to apply this to a temporal sequence of events, using our proposed definition of blame. The latter is particularly important in a reinforcement learning setting, where the policy must be trained from data and therefore the effect of the policy, i.e. which of the actual causes are actually under control, cannot be known beforehand. 
	
	Our definition of blame rests on a criteria of whether the time an event occurs can be hastened or postponed. This view of blame is quite rudimentary compared to views found in philosophy (c.f. \cite{hertzberg1975blame,kagan1989limits}. Elsewhere, blame has been defined as a trade-off that involves the cost to the agent~\cite{halpern2018towards}. The trade-off between costs and blame will factor into our method in the reward function \cref{eq:reward0}. In other words, the agent will \emph{in its choice of actions} factor in the cost, but not in its definition of being blameworthy.

	Our method ultimately learns by using a modified reward signal, so it is natural to ask if \emph{"reward is enough"}~\cite{SILVER2021103535}. We don't claim that morality challenges whether reward-maximization \emph{is the best way to train agents}, but rather that \emph{if} we want the agent to take blame or responsibility into account, the modification to the reward should be based on its formal definition. This is what we have attempted to do in our method.
	
	In experiments, we have seen that our method will learn which variables it can influence and thereby use a reward function that conforms to the notion of morality that we have assumed to be correct: the agent reliably learns a more costly policy, but one where it is not to blame when the forest burns down. Our experiments confirmed that this occurs through a correct assignment of blame.

	Although we have dubbed our method moral reinforcement learning, we are obviously not making a claim that our simple criteria of avoiding being the cause of a bad outcome captures all morality. An obvious option for future work is to consider \emph{intent}, which has been given a formal treatment in actual causality\cite{halpern2018towards}. It would be interesting to examine applications of this definition to situations where intent has been singled out as discriminating between good and bad behavior~\cite{kagan1989limits}.
	
	To the extent that causal moral notions such as blame can be specified using SCMs, we believe that our approach illustrates a plausible way to integrate them with reinforcement learning, thereby allowing machine-learning models to learn and act using these moral intuitions. This opens up a possibility both to create machines that are more moral, and also to concretely see what actual behaviors philosophical formulations of moral principles will lead to when implemented.

	\bibliographystyle{unsrt}
	\bibliography{references}
\end{document}